\documentclass[a4paper, conference]{ieeeconf}

\pdfoutput=1
\IEEEoverridecommandlockouts
\usepackage[bookmarks=false]{hyperref}
\usepackage{graphicx} 
\usepackage{subfigure}
\usepackage{xcolor}
\usepackage{multirow}
\usepackage{siunitx}
\usepackage{cleveref}
\usepackage{fullwidth}

\title{Compositional Attention Networks for Interpretability in Natural Language Question Answering}

\author{
Muru Selvakumar$^1$ \hspace{1em}
Suriyadeepan Ramamoorthy$^2$ \hspace{1em}
Vaidheeswaran Archana$^3$ \hspace{1em}
Malaikannan Sankarasubbu$^4$ \\
\{$^1$smurugan, $^2$suriyadeepan.ramamoorthy,  $^3$a.iyer, $^4$malaikannan.sankarasubbu\}@saama.com
\\  Saama AI Research \\
    Chennai, India  
}

\begin{document}

\maketitle

\thispagestyle{empty}
\pagestyle{empty}

\begin{abstract}
MAC Net \cite{macnet} is a compositional attention network designed for Visual Question Answering. We propose a modified MAC net architecture for Natural Language Question Answering. Question Answering typically requires Language Understanding and multi-step Reasoning. MAC net's unique architecture - the separation between memory and control, facilitates data-driven iterative reasoning. This makes it an ideal candidate for solving tasks that involve logical reasoning. Our experiments with 20 bAbI tasks, demonstrate the value of MAC net as a data-efficient and interpretable architecture for Natural Language Question Answering. The transparent nature of MAC net provides a highly granular view of the reasoning steps taken by the network in answering a query.
\end{abstract}

\section{Introduction}
There is a growing interest in the Machine Learning community to build explainable Artificial Intelligence. Research in Interpretability \cite{explanation} focuses on explaining decisions made by machines to a human observer. A model is more interpretable than another if it's decision making process is more transparent than the other. Transparency, in this case, is relative to the human observer. Can a human subject comprehend the decision-making process? Hence research in interpretability draws inspiration from the fields of  philosophy, psychology, and cognitive science, to define and evaluate explanations, considering the cultural and socially-conditioned cognitive biases inherent in human beings \cite{rigorous}.

For a certain real-world problems, it isn't enough to arrive at an answer. An explanation as to how the model came to such an answer, is crucial. In those cases, it is clear that predicting an answer only partially solves the problem. Incompleteness in problem formulation necessitates interpretability. This under-specification of real world problems leads to learned biases in the model leading to gender-based or racial discrimination. A well-specified problem would include traits such as fairness, inclusivity, safety, etc,. 

Interpretability could be used as a tool to identify these biases in order to eliminate them. Concepts like gender, race, color, etc, are fairly high-level abstract "human" concepts. Our machine learning model is itself a source of knowledge. It learns these abstract concepts from the data. An interesting area of research in interpretability, is to identify the learned concepts in the model and measure the biases in them. For example, does the concept \textit{gender} $G$, of an applicant influence their likelihood of being selected $p(selection | \theta, G) \ne p(selection | \theta)$ ? Yes? Then, the model is biased.

Most modern machine learning models, although powerful and highly accurate, are essentially black boxes. Understanding the reasoning behind a prediction is crucial to trusting \cite{whytrustyou} a model and deploying it in sensitive environments. Our work focuses on transparency of the prediction process which involves multi-step reasoning. By visualizing every reasoning step performed by our model, we are able to track the causal chain of reasoning starting with searching for initial piece of evidence, conditioned on the query, to arriving at the final answer. Thus, our model, in addition to answer the "what" question, also provides the logical steps taken in order to arrive at an answer, thus answering the "how" question.

\cite{molnar} has outlined different criteria to classify methods for machine learning interpretability. The model we propose is internpretable by design, hence it is intrinsically interpretable and it falls under the category of Model-specific interpretability. Our model provides explanations based on attention distributions extracted from model internals rather than studying relationship between the prediction and the features.

\subsection{Data}
bAbI \cite{babi} is a synthetic RC dataset, created by facebook researchers in 2015. The term synthetic data refers to data that is not extracted from a book or from the internet, but is generated by using a few rules that simulate natural language. This characteristic of bAbI places the weight of the task on the reasoning module rather than the understanding module.Question Answering data sets provide synthetic tasks for the goal of helping to develop learning algorithms for understanding and reasoning. 
AI's futuristic objective is that to create a dialogue platform automatically meanwhile, Question Answering data holds an important key in understanding dialogue. The way it works is the data's intent is to design different kinds of question sets which will further act as their skill set. This system typically replicates a software testing system. It has 20 different tasks of various reasoning patterns. Firstly it has a Single Supporting act, wherein the answer is usually a 'single word'. For example "John went to the gym. Where is John?". Then it moves on to harder tasks like to Two or Three Supporting Act. But the goal of such a data set is challenged while recognizing objects, negation and reasoning. For example those tasks with indefinite knowledge like " Mary might like chocolate or vanilla" such questions are answered with a "maybe" when asked about either. However the other sorts of simpler tasks the dataset does include is that of Yes/No Based or that of Counting of Lists/Sets.

\begin{table}[h!]
\begin{center}
\caption{bAbI Tasks.}
\label{tab:tasks}
\begin{tabular}{l|l}
\hline
single-supporting-act & basic-coreference\\
two-supporting-facts  & conjuction\\
three-supporting-facts & compound-coreference \\
twoarg-relations  & time-reasoning\\
three-arg-relations & basic-deduction\\
yes-no-questions & basic-induction\\
counting & positional-reasoning\\
list-sets & size-reasoning\\
simple-negation & path-finding\\
indefinite-knowledge & agent-motivations\\
\end{tabular}
\end{center}
\end{table}

\subsection{ Reasoning }
Babi Question Answering Dataset also includes slightly complex tasks of Reasoning. Informally there are three types of reasoning.  Abduction, Deduction, Induction. Induction tries to figure out the rules based on observed examples. Abduction can be thought of reverse engineering of events. It involves determining the premise or series of events which to led to something. But other types of reasoning reviewed in, include those of Positioning and Spatial Reasoning or Path Finding.For example you have a question that reads " Canada is present to the north of the USA", so when a question like "How do you go from USA to Canada?" the answer to it is the "north". Reasoning is mostly used to mean, deductive reasoning. Given set of premises and rules, deduction tests the validity of the conclusion. 

\section{ Model }
\subsection{Memory Net}
End-to-End Memory Networks \cite{memnet},consists of memory slots for representing each fact. It uses an iterative attention mechanism to score the importance of each fact and reduce them into a final joint representation of the question and facts.The sentences are then encoded into memory vectors by a simple embedding look-up based on the embedding matrix A. The question is encoded into the initial internal memory state, using the embedding matrix B. Attention mechanism is applied on the memory vectors conditioned on to produce attention weights. Another set of memory vectors are created using the embedding matrix C. An output memory representation 0 is produced by a weighted sum .A memory network is a model that iteratively applies attention mechanism on distributed representation of each fact in the context, to arrive at a representation of context, conditioned on the query.We then,select the most probably answer from a candidate set,conditioned on the final context representation and the query representation.The memory network solves 16\/20 tasks (>95\% accuracy)
 
\subsection{ MACNet}
A given story forms the basis for the model.The model used takes into account two things, a story and a question which is to be answered by the model. The input information is first encoded into a vector form before the reasoning operation begin. MACnet\cite{macnet} takes its inspiration from computer architecture, where there is a clear distinction between control and memory. It also closely resembles the computer architecture in its implementation.  

\begin{itemize}
    \item Control unit which figures out what kind of reasoning has to be done
    \item Read Unit which based on the reasoning instruction, reads relevant information from the knowledge base
    \item Write Unit which produces a memory vector based on the reasoning instruction and information extracted by the Read Unit.
\end{itemize}

The input module in our case is an LSTM \cite{lstm} network which encodes both the story and question. Output of the model is a probability distribution over all the answer words. The usual classifier network acts as the output module. The MACCell\cite{macnet} is sandwiched between these two. The number of steps of reasoning here is hyper-parameter. This is similar to number of hops in memory networks[CITE]. Each MACCell consists of three modules.

We describe the three modules of the MACCell from the original MACnet paper\cite{macnet} and then we contrast the difference between our network. 

Model section explain all three modules in detail. The primary advantage of MACCell is that, they can be cascaded to perform series of reasoning operations to achieve the desired result.

Like Recurrent Neural Networks, MACNet\cite{macnet} for its sequential reasoning ability, maintains a state $(c_i,  m_i)$ and is composed of control vector $c_i$, memory vector $m_i$. The initial state of the cell, $m_0, c_0$ is initialized with zeros.

We produce the $q_i$ input via linear projection, $ q_i = W^{d\times2d}[q_{i-1}; m_{i-1}] + b^d$. Where, $q_0$ is the final hidden state of the question input module.

\begin{equation}
  q_i = W^dq_{i-1} + b^d
\end{equation}

\subsubsection{Control Unit}
\begin{figure}[t!]
\centering
\includegraphics[scale=0.09]{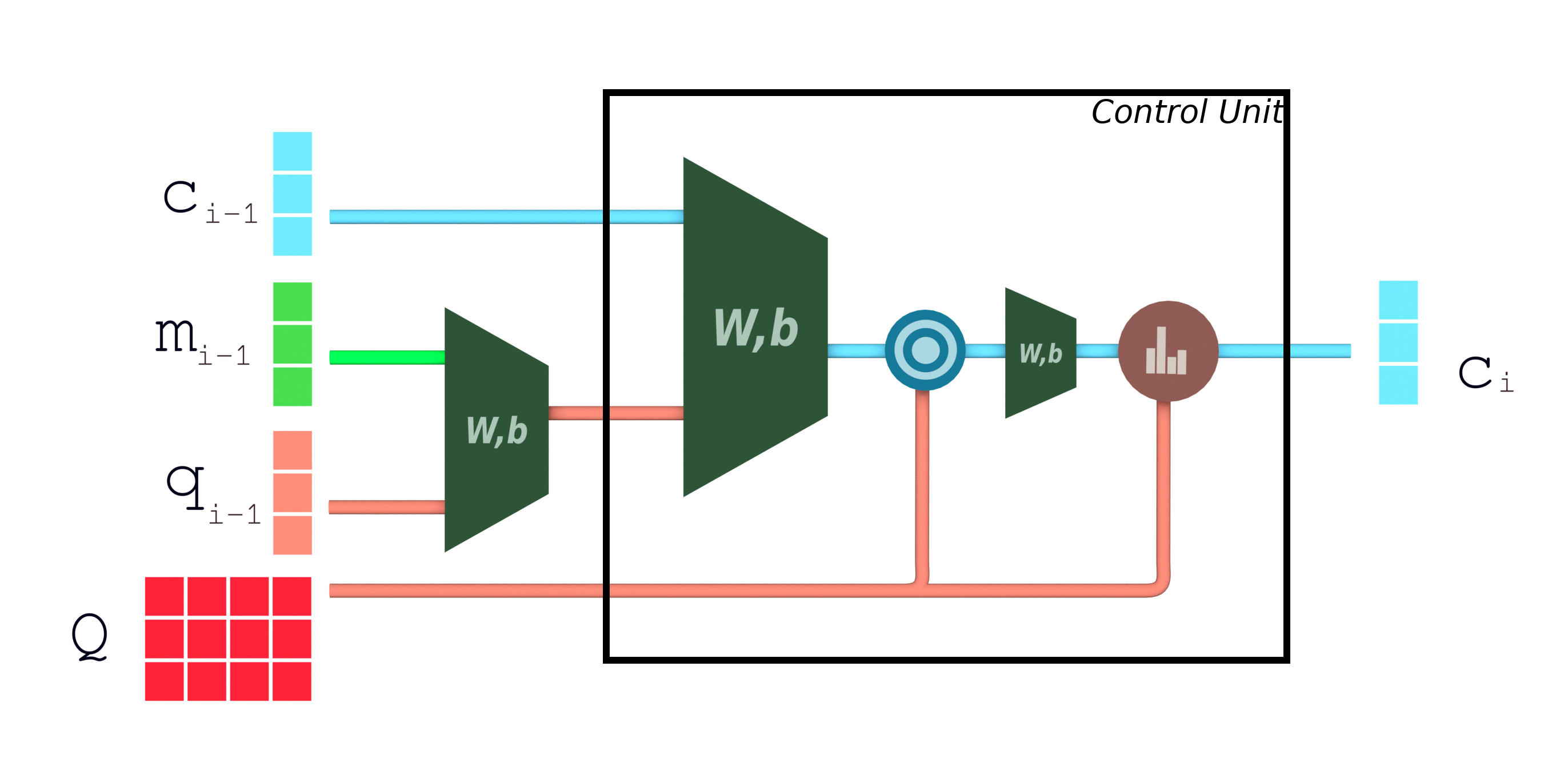}
\caption{Control Unit: The control unit produces the control vector which is weighted combination of hidden states of words in the question. The weighing is accomplished by the attention mechanism.}
\end{figure}

The control vector $c_i$ is the reasoning operation, for the $i^{th}$ step. In ADD:MACNet paper\cite{macnet}, Control unit takes in previous control vector $c_{i-1}$ and $q_i$ produces a new control vector $c_i$. We also feed previous memory vector $m_{i-1}$ to the control unit. This is to help the control unit, to have relevant information from not just the question but also from the story.

The control, question, memory vectors are linearly projected to produce intermediate control vector which is then used to weigh the hidden states of words in question $cw_0...cw_{L_q}$ based on similarity.
\begin{equation}
  cq_i = W^{d\times3d}[c_{i-1}; q_i; m_{i-1}] + b^d 
\end{equation}
\begin{equation}  
  ca_{i,u} = W^{1\times d}(cq_i \cdot  cw_u) + b^1
\end{equation}

The similarity/attention vectors $ca_{i,u}$ are transformed to create a attention distribution $cv_{i,u}$over the words in question. The sum of  $cw_0...cw_{L_q}$ by the attention distribution results in the control vector $c_i$
\begin{equation}  
  cv_{i,u} = softmax(ca_{i,u})
\end{equation}
\begin{equation}
  ci = \sum_{u=1}^{L_q} cv_{i,u}\cdot cw_u
 \end{equation}

\subsubsection{Read Unit}
\begin{figure}[t!]
\centering
  \includegraphics[scale=0.09]{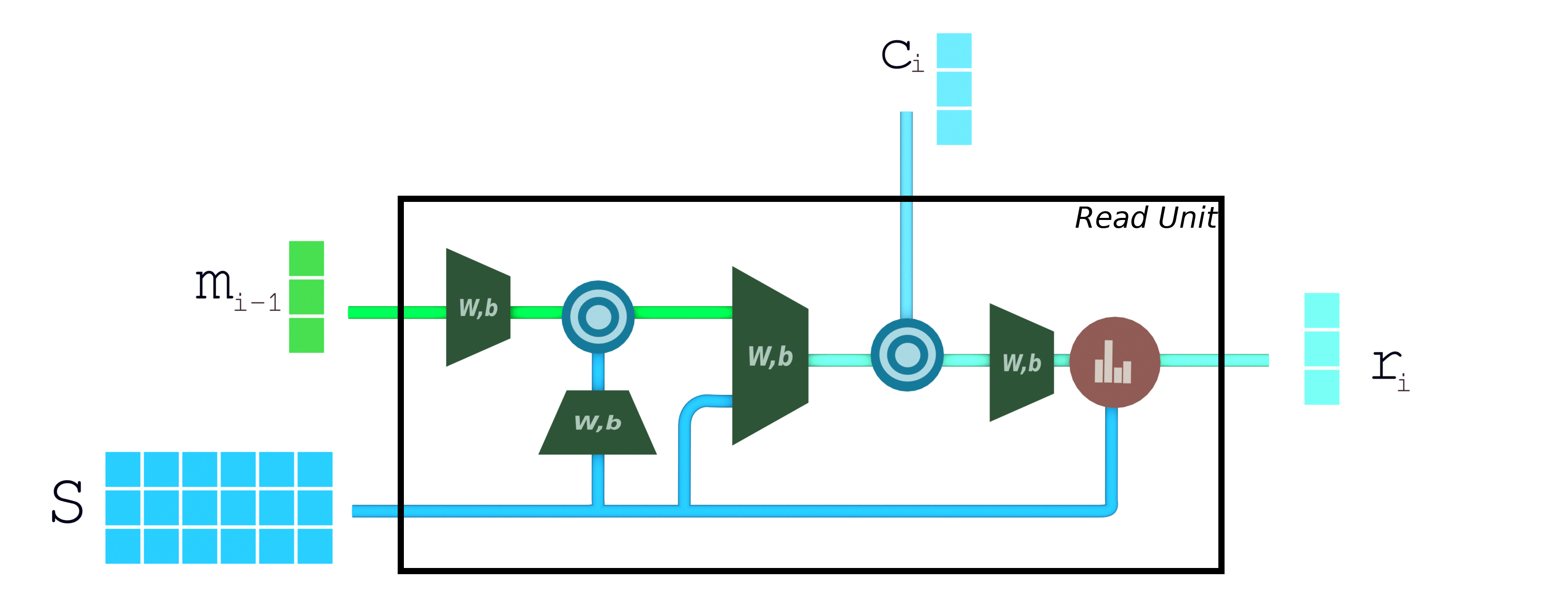}
  \caption{Read Unit: The read unit extract necessary information from the knowledge base for the current reasoning operation which is related to previously obtained memory.}
\end{figure}

The read unit, gathers relevant information from the knowledge base in our case the story, based on the current reasoning operation and previously obtained memory. This is a two step attention process, first one is to extract information based on the previously obtained memory, and the second to actually perform the reasoning operation.

The hidden states of the story and memory vector are projected into space for interaction. The memory vector is used as lens over the story to pickup relevant information based which might help the current reasoning step.
\begin{equation}
  S' = W^{d \times d}S + b^d 
\end{equation}

\begin{equation}
  m_{i-1}' = W^{d \times d}m_{i-1} + b^d 
\end{equation}

\begin{equation}  
  I_{i,v} = W^{1\times d}(S' + m_{i-1}') + b^1
\end{equation}

The interaction vector is then concatenated to hidden states of story and projected on to another space where they can be combined with control vector $c_i$ to produce an attention distribution $rv_{i,v}$. 
\begin{equation}  
  I_{i,v}' = W^{1\times d}[I_{i,v} ; S] + b^1
\end{equation}

\begin{equation}
  ra_{i,v} = W^{d \times d}(c_i \cdot I_{i,v}' ) + b^d
\end{equation}

\begin{equation}
  rv_{i,v} = softmax(ra_{i,v})
\end{equation}
Like in Control unit, this attention distribution is then used to produce a weighted sum of story vectory $r_i$
\begin{equation}
  r_i = \sum_{v=1}^{L_s} rv_{i,v}\cdot cw_v
 \end{equation}

\subsubsection{Write Unit}
\begin{figure}[h]
\centering
  \includegraphics[scale=0.09]{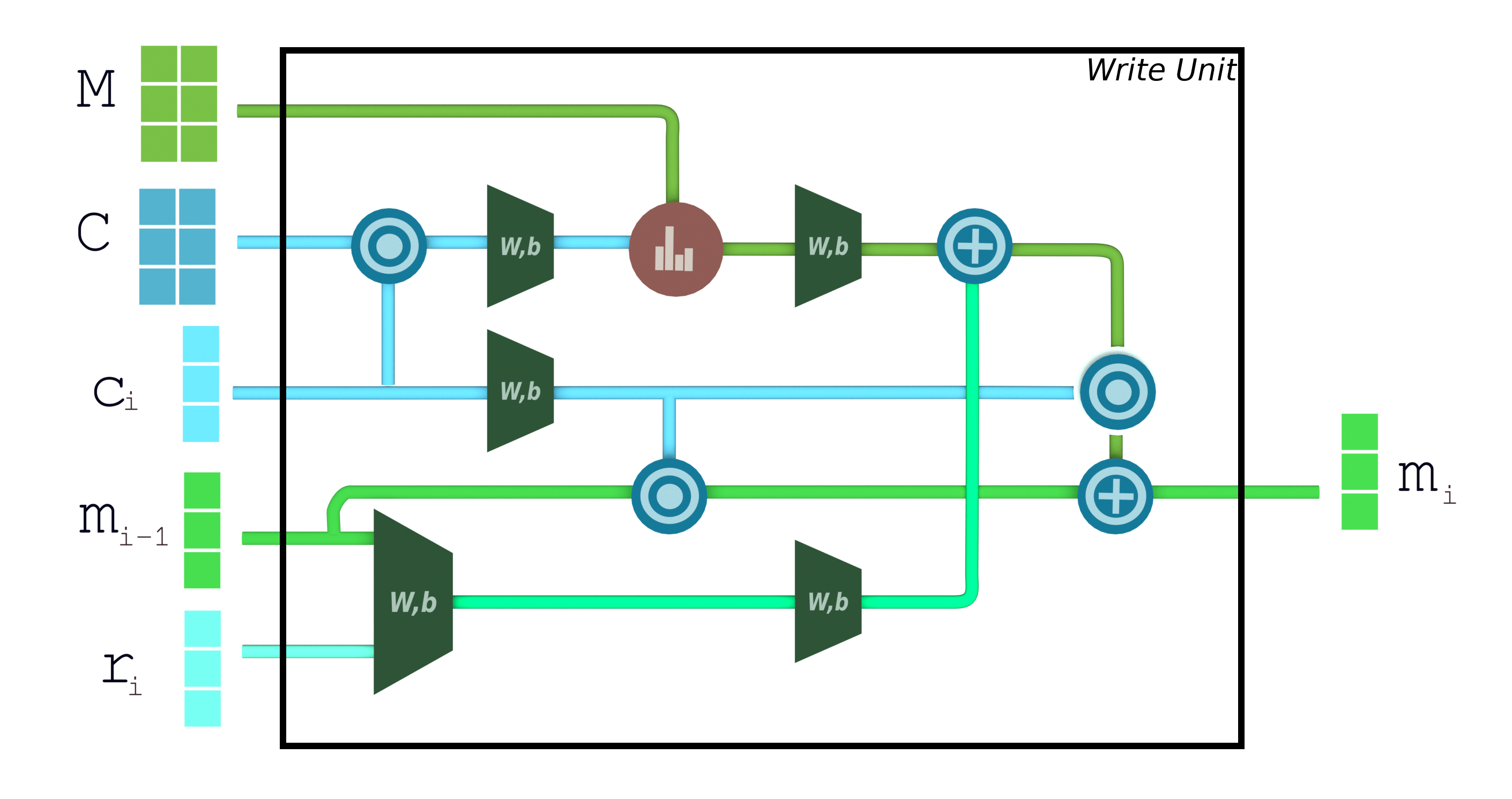}
  \caption{Write Unit: The write unit integrates the information retrieved from the knowledge base into the recurrent memory state, producing a new intermediate result $m_i$ that corresponds to the reasoning operation $c_i$.
  }
\end{figure}
Write Unit: The write unit integrates the information retrieved from the knowledge base into the recurrent memory state, producing a new intermediate result $m_i$ that corresponds to the reasoning operation $c_i$

The Write unit, in its bare essentials, produces a vector, $m_i^{info}$ that combined information from previous memory vector, $m_{i-1}$ and information gathered, $r_i$ by read unit.
\begin{equation}
  m_i^{info} = W^{d \times 2d}[m_{i-1}; r_i] + b^d 
\end{equation}

But not all tasks require same amount of reasoning. Simpler questions, can be answer quickly without having to do $p$ many number of reasoning operation. In such cases, it helps to conduct the previous memory $m_{i-1}$ to the next step to bypass the complexity. The control vector $c_i$ is squashed to a scalar which is then used to produce weighted sum of previous and current memory vector. $m_i$ = $m_i^{info}$
\begin{equation}
  c_{i}' = W^{1 \times d}c_i + b^1
\end{equation}
\begin{equation}
  m_i = \sigma(c_i')\cdot m_{i-1} + (1-\sigma(c_i)) \cdot m_i
\end{equation}

On the other hand, some tasks require more complicated reasoning patterns. By nature, MACNet is able to perform sequential reasoning. But some tasks might demand, tree-like or graph-like reasoning patterns. For such cases, we exploit, all the previous control vectors $C = (c_0 ... c_{i-1})$ and memory vectors $M = (m_0 ... m_{i-1})$. The control vectors $C$ is weighted by the current control vector $c_i$,
\begin{equation}
  sa_{ij}' = softmax(W^{1 \times d}[c_i \cdot C^{d \times i-1}] + b^1 )
\end{equation}
 and the resulting vectors are used to produce a weighted combination of the previous memories $M$. 
\begin{equation}  
  m_{i}^{sa} = \sum_{j=1}^{i-1}(sa_{ij} \cdot m_j)
\end{equation}
\begin{equation}
  m_{i}' = W^{d \times d}m_i^{sa} + W^{d \times d}m_i^{info} + b^d
\end{equation}

\subsubsection{Models Interpretability}

 
\begin{figure}
\includegraphics[scale=0.8]{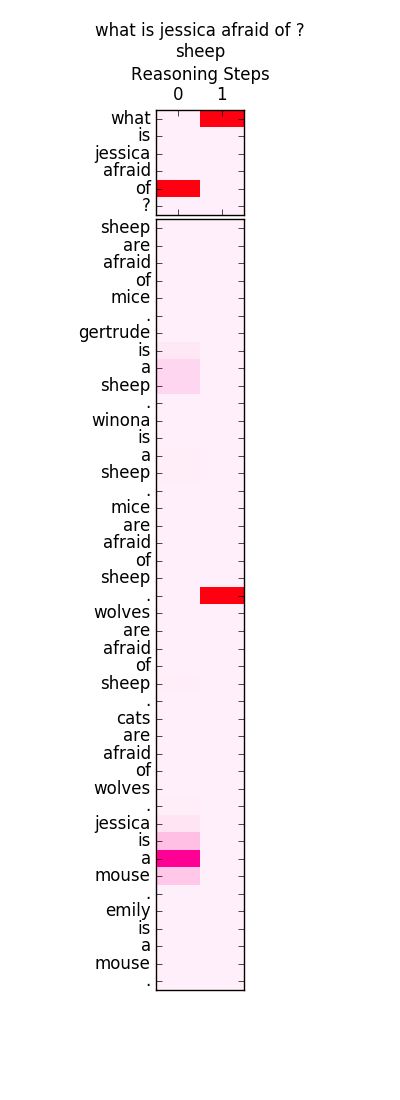}
\caption{The model first tries to find what Jessica is, when it finds out that Jessica is a mouse and then it goes onto figure what mice are afraid of.}
\label{fig:jessica}
\end{figure}

Macnet's modular design takes inspiration from Computer Architecture. From the previous sections, it is clear that there is an explicit separation between control and memory. Answering complex natural language questions based off a natural language text, requires Natural Language Understanding and Reasoning. A typical neural network model is not modular (i.e.) there is no clear separation between understanding and reasoning modules. This makes it hard and in most cases, impossible to interpret the dynamics of the model. This lack of separation between control and memory, also limits the dynamical behaviour of the model. 

Most Question Answering tasks require the model to adopt different behaviours for inferring answers to different queries. LSTM-based Recurrent Neural Networks \cite{lstm} by nature of their architecture, perform sequential reasoning by default. There are queries that require tree-like reasoning. This requires a different mode of operation. By separating the control and memory units, Macnet is able to dynamically adapt to the query and perform different kinds of reasoning.

\begin{figure*}[h]
  \centering
    \includegraphics[width=0.9\linewidth]{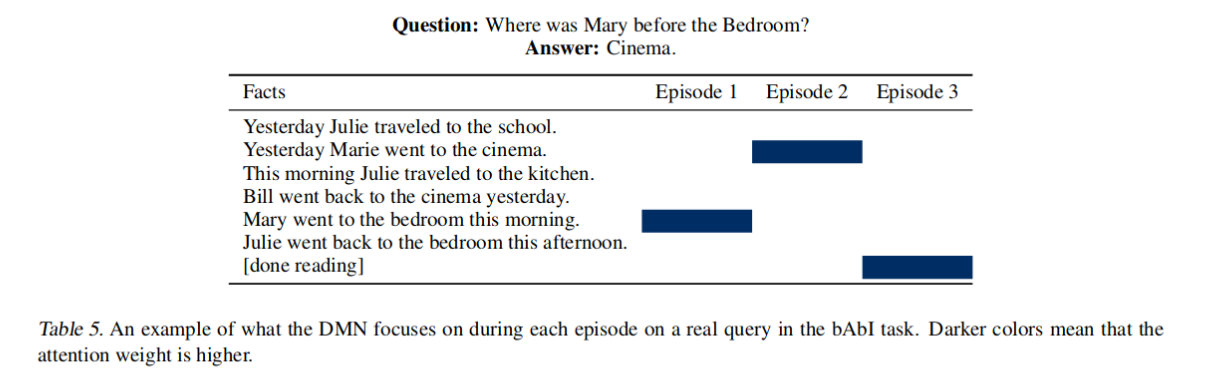} 
    \caption{From Ankit Kumar, et al. (DMN) \cite{dmn} here}
\label{fig:subim1}
\end{figure*}

This also puts Macnet at an advantage when it comes to Interpretability, as the control/memory separation improves model interpretability. Multiple control and memory representations are created iteratively by control and read/write units. Representations at any step i,are dependent on previous representations. These dependencies are summarised in attention distributions created during each step. By following the flow of attention along the time-steps, we could observe the reasoning pattern adapted by the network to answer the query.

Apart from the dependencies between representations at different time-steps, we could also observe the attention distribution across words of story and query (hidden states). This provides us a more granular view of what is in focus during each step. Most queries are compositional and require compositional reasoning to answer them. A query could be broken down into multiple smaller queries which require multiple steps to answer. By observing the attention distribution at a reasoning step, we could identify which sub-query is being answering in that step. The key goal of desigining an interpretable system is to answer the question "why"? Although Macnet doesn't answer this question directly, the in-built transparency enables us to understand the network behaviour by following the breadcrumbs of attention. 

\subsection{ Training}
The samples from different sub-tasks are mixed in a batch and fed to the training. The batch size is 32. The MACNet state size is 40 and LSTM state size is 20. 

\section {Results and Analysis}
\subsection{Results of the model}

\begin{table}[]
\centering
\begin{tabular}{||c c||}
\hline
Task & Accuracy \\ [0.5ex] 
\hline\hline
Single Supporting Fact& 100\% \\ 
\hline\hline
Two Supporting Facts& 91\% \\
\hline \hline
Three Supporting Facts& 87\% \\
\hline \hline
Two Argument Relations& 100\& \\
\hline \hline
Three Argument Relations& 93\& \\
\hline \hline
Yes-No Questions& 100\% \\
\hline \hline
Counting& 66\% \\
\hline \hline
List-Sets& 97\% \\
\hline \hline
Simple Negation& 100\% \\
\hline \hline
Indefinite-knowledge& 100\% \\
\hline \hline
basic coreference& 100\% \\
\hline \hline
conjunction& 100\% \\
\hline \hline
compound coreference& 100\% \\
\hline \hline
time reasoning& 93\% \\
\hline \hline
basic-deductions& 100\% \\
\hline \hline
basic-inductions& 46\% \\
\hline \hline
positional-reasoning& 74\% \\
\hline \hline
size reasoning& 78\% \\
\hline \hline
path-finding& 33\% \\
\hline \hline
agent-motivations& 96\% \\
\hline \hline
main& 92\% \\
\hline \hline
\end{tabular}
\end{table}


\subsubsection{Interpretable Results}

A query "What is Jessica afraid of?", based on the context (figure \ref{fig:jessica}), can be answered in two steps - 1. What is Jessica? Jessica is a mouse 2. What are mice afraid of? Mice are afraid of the inevitability of impending doom. Based on the attention distribution (from figure 4), we know that the network focuses on the "What is Jessica?" during the first step and then in the next step, once the network figures out the answer to the first sub-query as "mouse", it focuses on "What are mice afraid of?". This resembles the compositional reasoning adopted by rational beings in answering complex (compositional) queries. 
 
Consider the context and query in figure \ref{item:story_again}. "What is daniel carrying?". Notice that there are separate attention distributions on the query and the story. The network identifies "what" as the keyword to search the story. The search happens in multiple steps. The network identifies football and milk as objects that suit the query "what", in different steps. The attention distribution on the story is tied to the attention distribution on the query. This allows the network to select a keyword in the query, corresponding to a sub-query and condition the inference on the story on that keyword. This combined view of evidence in story and keywords in the query, provides us a clear view of the network's behaviour.

\subsection{Analysis}
\begin{figure}[h]
\includegraphics[scale=0.5]{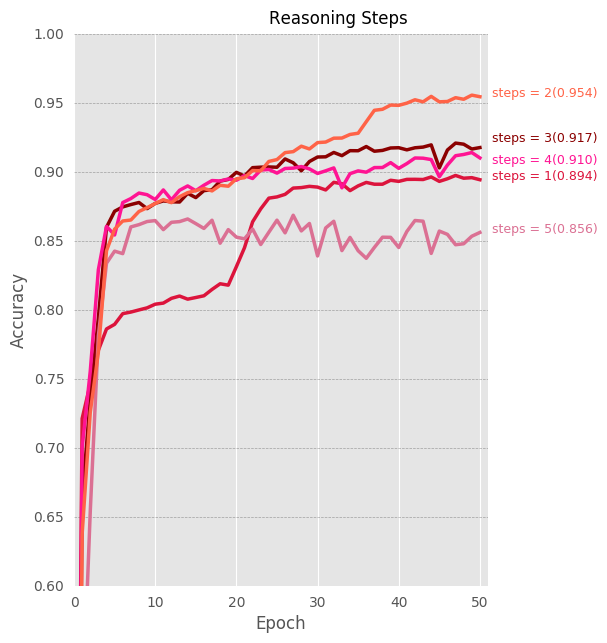}
\caption{The figure, shows the performance of model with reasoning steps from 1 upto 5. Even though models with more reasoning steps perform well at first, they start deteriorating soon.}
\end{figure}

We study the influence of number of reasoning steps. The figure, shows the performance of model with reasoning steps from 1 upto 5. Even though models with more reasoning steps perform well at first, they start deteriorating soon. The complexity bAbI dataset\cite{babi}, is well aligned with two step reasoning. It is appears questions in bAbI dataset\cite{babi}, do not demand reasoning steps beyond two.
 
\subsection{Ablation Study}

\begin{enumerate}
\item \label{item:diff_lstm} Different LSTM\cite{lstm} to encode story and question: Using different LSTM\cite{lstm} for encoding story and question results in 1 percent drop in accuracy.
  
\item \label{item:story_again} Without story again in Read Unit: In Read Unit we extract information relevant from story based on the previously retrieved memory. Then we concatenate the story with the extract. This enable the unit to extract information w.r.t to current control word, which is independent previous reasoning operation. Figure \ref{fig:with_story_again} and \ref{fig:without_story_again} contrasts the difference in operation.
  
\item \label{item:prev_mem} Without using previous memory in Control Unit: by preventing the Control Unit from accessing previous memory, we restrict the model to rely on just the control information from previous step to generate new control. We see steep drop in performance, because the model is not able to use any new information from the context. 

\item \label{item:graph_reasoning} Without graph reasoning: by passing the graph reasoning, i.e exploiting the previous memory words based on weighted combination of control words, also results in performance drop. 

\begin{figure}[h]
\centering
\includegraphics[scale=0.5]{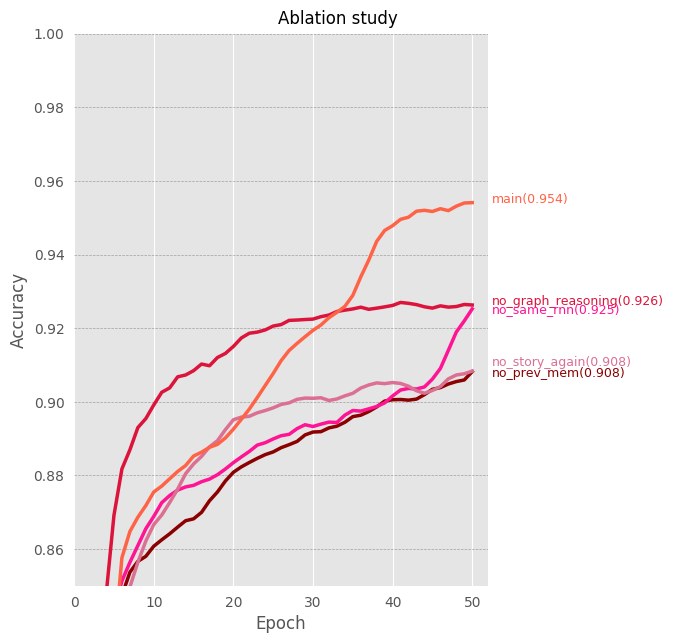}
\caption{Ablation study: how different components affect the performance of the model}
\label{fig:ablation_study}
\end{figure}
\end{enumerate}

\subsection{Dataset Size}

\begin{figure}[h]
\centering
\includegraphics[scale=0.5]{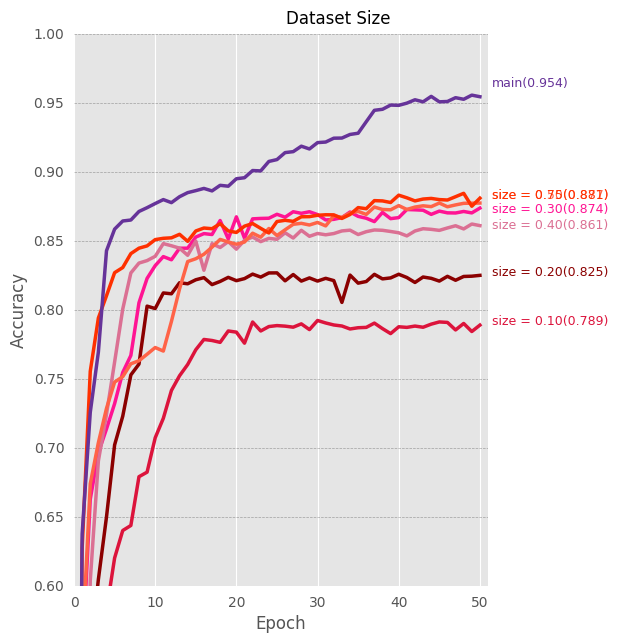}
\caption{With 10.0 percent of the dataset used for training, model achieves 79.4\% accuracy.}
\label{fig:dataset_subset}
\end{figure}

\begin{figure}[h]
  \begin{minipage}[b]{0.4\textwidth}
    \includegraphics[width=\textwidth]{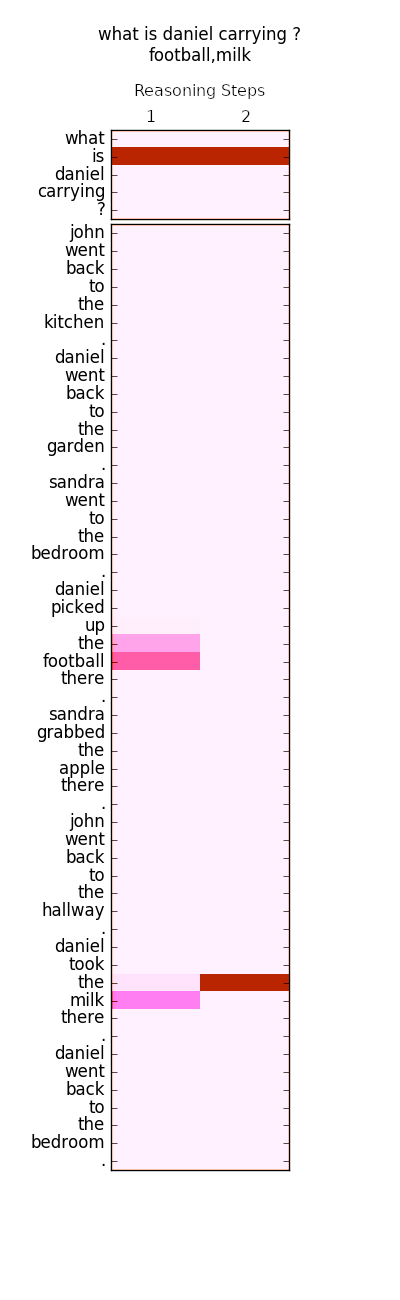}
    \caption{Re-reading  the  story  at  every  step helps to combine information which is irrelevant to previous reasoning operation, which is reflected in the plot above.}
    \label{fig:with_story_again}
  \end{minipage}
  \hfill
  \begin{minipage}[b]{0.4\textwidth}
    \includegraphics[width=\textwidth]{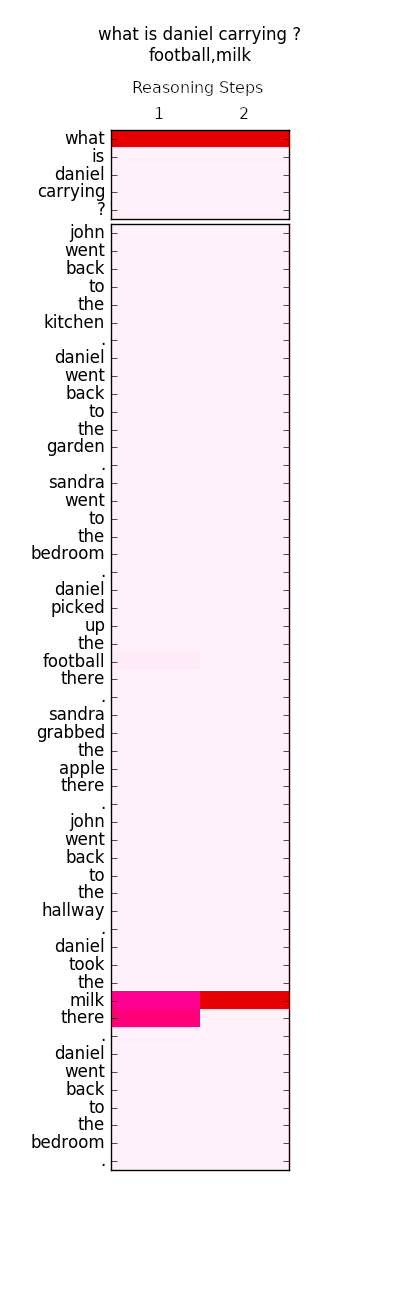}
    \caption{Though the model is able to answer correctly, how the answer is extracted is relatively opaque}
    \label{fig:without_story_again}
  \end{minipage}
\end{figure}

The figure \ref{fig:dataset_subset} shows the performance of model, when it is trained with varied subset (randomly sampled) of the dataset. We also tried training with shorter story lengths and test the model on lengthier stories. But this turned out to be redundant, by the story length distribution, sample size represented by stories with length less than 50, and 75 percent subset of the dataset is almost same.

\section {Conclusion}

The interesting observation in this work, is that the MACnet increased interpret-ability when it comes to deductive reasoning. The following examples show some of the samples from deductive reasoning task. See how the reasoning happens, step by step.

We propose a modified MAC net for natural language reasoning. We experiment with 20 bAbI tasks. Our results show that MAC net, designed for Visual Question Answering, can be adapted for Question Answering. We show that MAC net is capable of performing complex iterative reasoning tasks. In addition, the transparent nature of the network makes it possible to understand it's behaviour by observing the attention distribution across reasoning steps. By studying the attention distributions at each step, we could understand the model behaviour at a more granular level. These results show that MAC net which features superior interpretability, data-efficiency and dynamic iterative reasoning, is an ideal candidate for the complex task of Question Answering. 

 \onecolumn
\pagebreak
\twocolumn

\bibliographystyle{plain}
\bibliography{refs}


\end{document}